\documentclass[runningheads]{llncs}
\usepackage{enumerate}
\usepackage{url}
\usepackage[colorlinks=true, linkcolor=black, urlcolor=black, citecolor=black]{hyperref}

\usepackage[T1]{fontenc}
\usepackage{graphicx}
\usepackage{float}
\begin{document}
\title{RAG-MCP: Mitigating Prompt Bloat in LLM Tool Selection via Retrieval-Augmented Generation}
\titlerunning{RAG-MCP}
%
\author{Tiantian Gan\inst{1,2} \and Qiyao Sun\inst{1,2}}
\authorrunning{Gan and Sun}
\institute{Beijing University of Post and Communications, Beijing, China
\and 
Queen Mary University of London, London, UK\\
\email{jp2022213034@qmul.ac.uk, jp2022213402@qmul.ac.uk}
}
\maketitle              
\begin{abstract}
Large language models (LLMs) struggle to effectively utilize a growing number of external tools, such as those defined by the Model Context Protocol (MCP)\cite{IntroducingMCP}, due to prompt bloat and selection complexity. We introduce RAG-MCP, a Retrieval-Augmented Generation framework that overcomes this challenge by offloading tool discovery. RAG-MCP uses semantic retrieval to identify the most relevant MCP(s) for a given query from an external index before engaging the LLM. Only the selected tool descriptions are passed to the model, drastically reducing prompt size and simplifying decision-making. Experiments, including an MCP stress test, demonstrate RAG-MCP significantly cuts prompt tokens (e.g., by over 50\%) and more than triples tool selection accuracy (43.13\% vs 13.62\% baseline) on benchmark tasks. RAG-MCP enables scalable and accurate tool integration for LLMs.
\end{abstract}
\section{Introduction}
\subsection{ Background and Motivation}

Large Language Models (LLMs) have demonstrated remarkable capabilities in natural dialogue, reasoning, and even code generation. However, they remain fundamentally constrained by the knowledge encoded in their parameters and the fixed context window available at inference time. In essence, an LLM without external access is “trapped” with only its training data and cannot easily update its knowledge or perform actions in the world \cite{patil2024gorilla}. To address this limitation, recent efforts have focused on augmenting LLMs with \textbf{external tools and function-calling} abilities \cite{chen2024enhancing}. By invoking tools (e.g. web search, databases, calculators) via defined functions or APIs, an LLM can fetch up-to-date information and execute complex operations beyond its built-in repertoire \cite{patil2024gorilla}. This paradigm - often referred to as zero-shot tool use or function calling — allows AI assistants to interface with the latest data and services, unlocking applications from real-time knowledge queries to specialized tasks in finance and travel planning \cite{chen2024enhancing}. In fact, major AI providers have embraced this trend: for example, leading LLM platforms now support plugin APIs and structured function calls so that models like GPT-4 or Claude can invoke external services through well-defined interfaces \cite{patil2024gorilla}.

In the research community, a variety of approaches have been proposed to enable and improve LLM tool use. Prompt-based strategies such as \textbf{ReAct} intermix reasoning steps with action commands, allowing an LLM to decide when to consult a tool in the context of a multi-turn “thought process” \cite{ReAct}. Model-centric approaches have also emerged: for instance, \textbf{Toolformer} fine-tunes an LLM to autonomously decide which API to call, when to call it, and how to incorporate the result, given only a handful of demonstrations per tool \cite{NEURIPS2023_toolformer} Other researchers have improved tool-use by incorporating it into training data and model tuning. This includes blending function call demonstrations into instruction-following datasets and exploring prompt formats that effectively describe available functions to the model \cite{chen2024enhancing}. Such efforts have markedly enhanced zero-shot tool usage performance. For example, fine-tuning a model on API call tasks with extensive tool-use data can yield impressive results – the \textbf{Gorilla} system augmented a 7B LLaMA-based model with relevant API documentation retrieval, enabling it to outperform even GPT-4 in generating correct API calls for a wide range of tools \cite{patil2024gorilla}. An important insight from these works is that providing just-in-time relevant context (be it through optimized prompts or retrieved documentation) greatly boosts the accuracy of an LLM’s tool selection and use, while mechanisms for the model to explicitly decide on tool use (such as special decision tokens for “answer vs. act”) can further improve reliability \cite{chen2024enhancing}.

Despite this progress, a new challenge arises as we scale up the number of tools available to an LLM. Most prior studies and deployments consider a relatively small set of tools or APIs, often hand-picked and easy for the model to handle within a prompt \cite{patil2024gorilla}. In practice, however, the ecosystem of tools is rapidly expanding. For instance, Anthropic’s recently introduced \textbf{Model Context Protocol (MCP) }defines a universal, open standard for connecting AI systems with external data sources and services. MCP enables a single assistant to interface with many data repositories and business tools through a unified protocol, replacing fragmented one-off integrations. As a result, an advanced LLM agent could soon have dozens of functions at its disposal – from Google Drive and Slack connectors to GitHub, databases, maps, and more – all registered as MCP “tools” it can call \cite{IntroducingMCP}. This proliferation of available tools brings significant hurdles.

\textbf{Prompt Bloat} is one critical issue: providing the definitions or usage instructions for every possible tool in the model’s context would consume an enormous number of tokens and risk overwhelming the model. It has been observed that it is effectively impossible to describe a large collection of APIs or tools in a single prompt as their number grows, and many APIs have overlapping functionalities with only nuanced differences. Including too many at once not only exhausts the context length, but can also confuse the model – the functions may start to blur together. This leads directly to a second issue: \textbf{decision overhead}. With a long list of tools (many of them similar in scope), the model faces a more complex decision when choosing if and which tool to invoke. The greater the choice, the higher the chance of error, such as selecting an suboptimal tool or misinterpreting what a tool does. Indeed, even state-of-the-art models can misfire in such settings: for example, in a scenario with numerous API options, GPT-4 was reported to hallucinate an API that doesn’t actually exist, and Anthropic’s Claude picked the wrong library for the user’s request \cite{patil2024gorilla}. These failure cases underscore that \textbf{naively scaling up the toolset can degrade an LLM’s performance}, due to both the capacity strain on the prompt and the ambiguity in the model’s decision process.

\begin{figure}[H]
    \centering
    \includegraphics[width=0.8\textwidth]{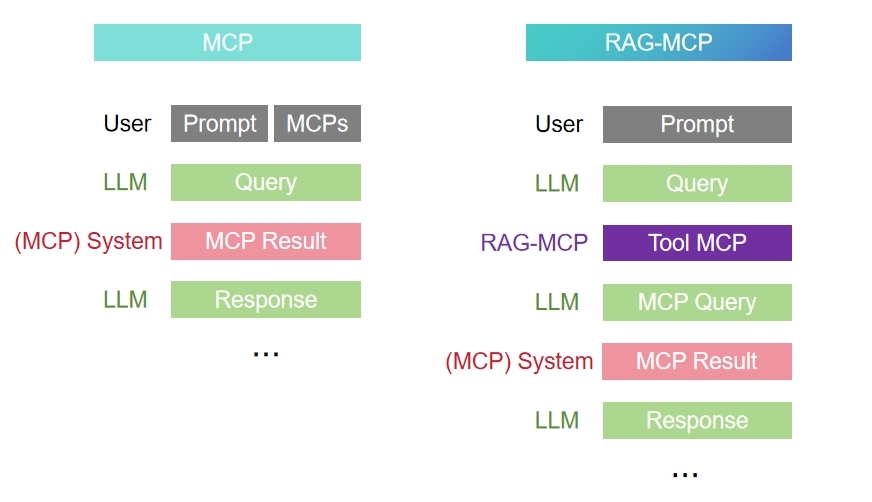} 
    \caption{Comparation between MCP and RAG-MCP during inference}
    \label{fig:process}
\end{figure}

To tackle these challenges, we propose \textbf{RAG-MCP}, a solution that marries Retrieval-Augmented Generation (RAG) with the Model Context Protocol framework. The key idea of RAG-MCP is to avoid presenting all tools to the language model at once, and instead dynamically retrieve a relevant subset of tools based on the user’s query. In our approach, the numerous available tool descriptions (MCP function schemas, usage examples, etc.) are stored in an external memory indexed by their semantics. When a new query arrives, a dedicated retriever (e.g. a vector-space semantic search) first selects the top-$k$ candidate tools that are most likely to be useful for that query. Only these $k$ tool descriptions are then injected into the LLM’s prompt (or provided via the function-calling API), greatly reducing context length and complexity. This retrieval step serves as a form of focused \textbf{context filtering}, which cuts down on prompt bloat and guides the model’s choice. The approach is analogous to how retrieval-augmented QA systems work: rather than feed the entire Wikipedia to the model, one retrieves only the relevant articles \cite{NEURIPS2020_6b493230_RAG}. Here, instead of static knowledge, we retrieve 
\textbf{actionable tool knowledge} on the fly. An added benefit is extensibility – because the tool information lives in an external index, new tools or updated APIs can be incorporated by updating that index without retraining the LLM, ensuring the system remains up-to-date \cite{patil2024gorilla}. In short, \textbf{retrieval helps tame the growing toolset} by providing the right tools at the right time, thereby reducing the model’s decision burden.

\subsection{Contributions} 
In summary, this paper makes the following contributions:
\begin{enumerate}[1. ]
\item \textbf{RAG-MCP Framework}: We introduce a novel architecture that integrates a retrieval mechanism with LLM function calling in the MCP setting. To our knowledge, this is one of the first frameworks to enable an LLM to handle a large arsenal of tools by querying a tool repository for relevant options instead of naively prompting with all tools. This design retains the flexibility of the open MCP ecosystem while imposing structure to maintain tractability.
\item \textbf{Scalable Tool Retrieval}: We develop a semantic tool retrieval module that represents each available tool’s description in a vector space and efficiently matches user queries to the most pertinent tools. This significantly reduces prompt size and complexity (mitigating prompt bloat) and improves decision making by narrowing the choices. The LLM, guided by this retrieved context, can more accurately select and use the correct external tool, even as the total number of tools grows large. Notably, our approach allows new tools to be added on the fly by indexing them, without requiring additional fine-tuning of the LLM.
\item \textbf{Improved Tool-Use Performance}: Through comprehensive experiments, we demonstrate that RAG-MCP effectively addresses the performance degradation that occurs with naively scaling up the tool set. On a suite of tool-augmented NLP tasks, we show that as the number of available functions increases, a baseline LLM’s success rate in selecting and executing the correct tool drops markedly (illustrating the aforementioned challenge). However, under the RAG-MCP strategy, the model’s performance is largely restored to its original level, and in some cases even exceeds the small-toolset baseline. In particular, RAG-MCP yields substantially higher accuracy in choosing the appropriate tool and reduces errors such as hallucinated or mis-parameterized function calls. These results underscore the efficacy of using retrieval to scale up tool-use: the proposed method enables an LLM to maintain high tool-selection accuracy and reliability even with a large pool of tools, paving the way for more scalable and capable tool-augmented AI systems.
\end{enumerate} 

Overall, our work demonstrates that the integration of retrieval-based context management is a promising direction to counteract the challenges of tool proliferation in LLMs. By enabling models to learn which tool to use out of many and only providing information for those tools, \textbf{RAG-MCP} offers a practical solution for the next generation of AI agents operating with extensive toolkits. It combines the strengths of retrieval augmentation and standardized tool APIs to ensure that more tools do not mean worse performance but rather a broader range of skills that the model can deploy accurately and efficiently.

\section{Related Work}
\label{relwork}
\subsection{Tool Use in LLMs}
LLMs have been augmented with external tools to overcome limitations in arithmetic, retrieval, and code execution. \textbf{Toolformer} demonstrates a self‑supervised method by which a model learns when and how to call APIs such as calculators or search engines, improving zero‑shot performance across tasks \cite{NEURIPS2023_toolformer}. \textbf{ReAct} interleaves chain‑of‑thought reasoning with action steps to interact with external environments (e.g., a Wikipedia API), yielding more interpretable and accurate multi‑step solutions \cite{ReAct}. \textbf{WebGPT} fine‑tunes GPT‑3 in a simulated browser environment, training it to navigate, search, and cite sources for long‑form Q\&A, reducing hallucinations via grounded retrieval \cite{nakano2022webgpt}. More recently, \textbf{ChatGPT Plugins} introduced a production plugin ecosystem, allowing ChatGPT to access up‑to‑date information and third‑party services in a controlled, safety‑oriented framework \cite{ChatGPTplugins}.

\subsection{Retrieval‑Augmented Generation}
Retrieval‑Augmented Generation (RAG) first combined parametric LLMs with non‑parametric memory in a dense vector index, retrieving relevant passages at inference time to improve knowledge‑intensive tasks \cite{NEURIPS2020_6b493230_RAG}. Subsequent work has extended RAG to broad NLP paradigms, including modular and advanced RAG variants that dynamically adapt retrieval per token or per query \cite{gao2023retrieval}. RAG’s decoupling of memory access and generation inspires our MCP‑RAG approach, wherein MCP discovery is treated as a retrieval subproblem, orthogonal to core text generation.

\subsection{Model Context Protocol}

The Model Context Protocol standardizes LLM‑to‑API interactions by bundling resource prompts, authentication, and parameter schemas into modular “MCP” servers. MCPs act as function‑call extensions, similar to OpenAI’s function‑calling API, but with greater community extensibility. The rapid growth of MCP repositories (4,400+ servers on mcp.so as of April 2025 \cite{MCPSO}) underscores the need for scalable discovery and validation mechanisms .

\section{Methodology}
Overview. We study how the number of available MCP servers affects an LLM’s ability to select and invoke the correct tool (“prompt bloat”) and present MCP‑RAG, a retrieval‑augmented framework that mitigates this degradation by dynamically retrieving only the most relevant MCP for each query.
\subsection{Prompt Bloat and the MCP Stress Test}
Modern LLMs must often choose among many possible external tools, each described by an MCP schema. As the count of MCPs grows, including all their descriptions in a single prompt leads to prompt bloat: the context window becomes saturated with distractors, reducing the model’s capacity to distinguish and recall the correct tool.

This phenomenon parallels the Needle‑in‑a‑Haystack (NIAH) test, which embeds a random fact (the “needle”) in the middle of a long context (the “haystack”) and measures an LLM’s ability to retrieve it under varying context lengths and depths \cite{NEURIPS2020_6b493230_RAG} \cite{ChatGPTfuctioncalling} . In NIAH, performance drops sharply as the haystack grows, revealing limits of in‑context retrieval.

Inspired by NIAH, we design an \textbf{MCP stress test} on WebSearch tasks: for each trial, we present the model with $N$ MCP schemas (one ground‑truth and $N - 1$ distractors) and ask it to select and invoke the correct WebSearch MCP. We vary $N$ from 1 to 11100 in 26 intervals, measuring selection accuracy, task success, prompt token usage, and latency. This setup quantifies how tool‑selection ability degrades with increasing MCP pool size.

\subsection{RAG-MCP Framework}
To overcome prompt bloat, RAG-MCP applies \textbf{Retrieval-Augmented Generation (RAG)} principles to tool selection. Instead of flooding the LLM with all MCP descriptions, we maintain an external vector index of all available MCP metadata. At query time:

\begin{enumerate}
    \item \textbf{Retrieval.} A lightweight LLM-based retriever (e.g., Qwen) encodes the user’s task description and performs a semantic search over the MCP index, returning the top-$k$ candidate MCPs most similar to the task \cite{NEURIPS2020_6b493230_RAG}.
    \item \textbf{Validation.} For each retrieved MCP, RAG-MCP can generate a few-shot example query and test its response to ensure basic compatibility, functioning as a “sanity check” before invocation.
    \item \textbf{Invocation.} Only the single best MCP description, including its tool-use parameters, is injected into the LLM prompt or function-calling API, which then performs planning and execution without concern for tool discovery \cite{NIVIDARAG}.
\end{enumerate}

This design yields several benefits:

\begin{itemize}
    \item \textbf{Reduced Prompt Size.} By supplying only relevant MCP metadata, RAG-MCP avoids context window overload even when the full tool registry is large.
    \item \textbf{Lower Cognitive Load.} The LLM no longer needs to sift through hundreds of distractors, improving selection accuracy and reducing hallucinations \cite{NIVIDARAG}.
    \item \textbf{Resource Efficiency.} Unlike conventional MCP clients (e.g., Claude or early GPT‑4 integrations) that must instantiate all registered MCP servers before interaction, MCP‑RAG activates only the selected MCP, lowering startup cost and enabling support for arbitrarily large toolsets without infrastructure bottlenecks \cite{ChatGPTfuctioncalling}.
    \item \textbf{Multi‑Turn Robustness.} In dialogues spanning many turns, the LLM need not re-include all MCP prompts; RAG-MCP’s retriever handles tool recall dynamically, freeing context space for task-specific reasoning.

\end{itemize}

\subsection{Three-Step Pipeline Diagram}
We summarize RAG-MCP’s operation in three core steps. The flowchart is shown in Fig. \ref{fig:process}:
\begin{enumerate}
    \item \textbf{Task Input → Retriever}: The user’s natural-language task is encoded and submitted to the retriever.
    \item \textbf{Retriever → MCP Selection \& Validation}: The retriever searches the vector index of MCP schemas, ranks candidates by semantic similarity, and optionally tests each via synthetic examples.
    \item \textbf{LLM Execution with Selected MCP}: The LLM receives only the selected MCP’s schema and parameters and executes the task via the function-calling interface.
\end{enumerate}

\begin{figure}[H]
    \centering
    \includegraphics[width=0.8\textwidth]{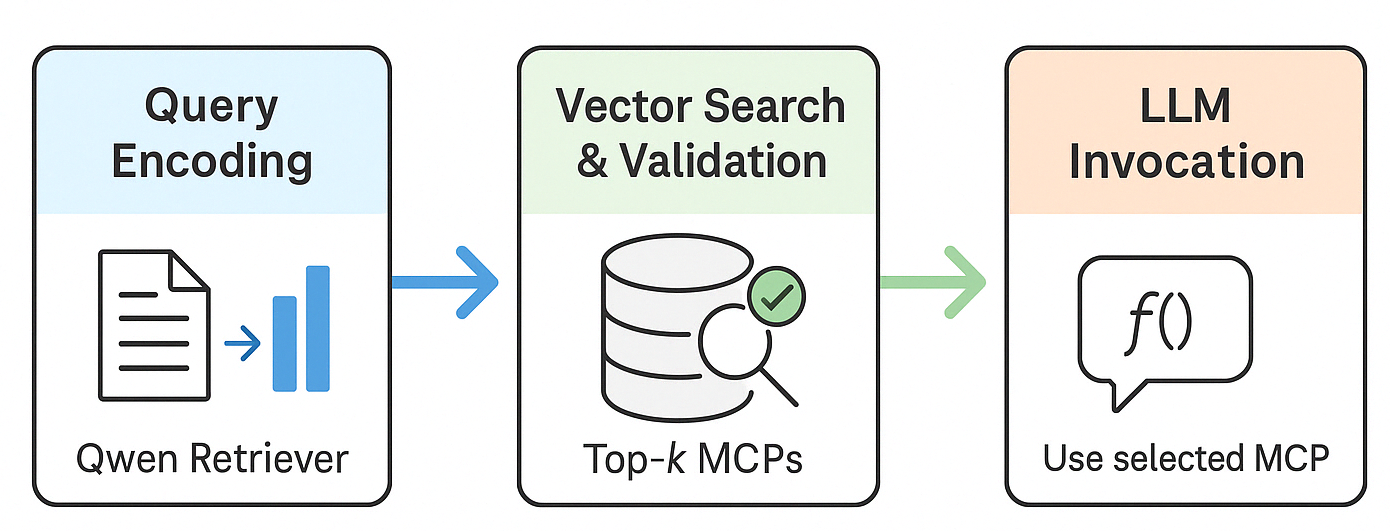} 
    \caption{RAG-MCP pipeline: (1) encode user query with Qwen-max, (2) re-trieve \& validate top-k MCPs, and (3) invoke chosen MCP}
    \label{fig:process}
\end{figure}

By decoupling tool discovery from generation, RAG-MCP ensures that LLMs can scale to hundreds or thousands of MCPs without suffering prompt bloat or decision fatigue, much as RAG systems avoid overwhelming an LLM with entire corpora by retrieving only relevant passages.

\subsection{Discussion}
Our methodology combines the rigor of \textbf{stress testing} (via the MCP stress test) with the effectiveness of retrieval-augmented tool use. The stress test quantifies the sharp performance drop that occurs when distractor MCPs swell the prompt, mirroring long-context recall failures in NIAH evaluations \cite{NIAH}. RAG-MCP then counteracts this by dynamically narrowing the toolset, reducing both prompt tokens and decision complexity, and thereby restoring—and often improving—task success rates.

Furthermore, by using an external index, RAG-MCP remains extensible: new MCPs can be added by indexing their metadata, without retraining the LLM. And by selectively activating servers on demand, it sidesteps the practical limits on simultaneous MCP instantiation faced by prior tool-augmented LLM deployments.

\section{Experiments}
\label{experiments}
\subsection{Stress Test}
\subsubsection{Setup} 

To quantify how an LLM’s tool‐selection ability scales with the size of the MCP pool, we conduct a stress test in which the number of candidate MCP servers, $N$, is varied from 1 to 11100 in intervals, while the key MCP server located from the top to the bottom. For each value of $N$, we randomly select one “ground‐truth” MCP (i.e., the only server capable of satisfying the task requirement) and $N - 1$ distractor MCPs drawn from our full registry of over 4,400 publicly listed servers \cite{MCPSO}. This design ensures that exactly one in every $N$ candidates is relevant. We then present the model with each of $20$ web‑search tasks, requiring it to (a) choose the correct MCP, (b) issue a valid query or answer, and (c) return the final result.

\begin{figure}[H]
    \centering
    \includegraphics[width=1.3\textwidth]{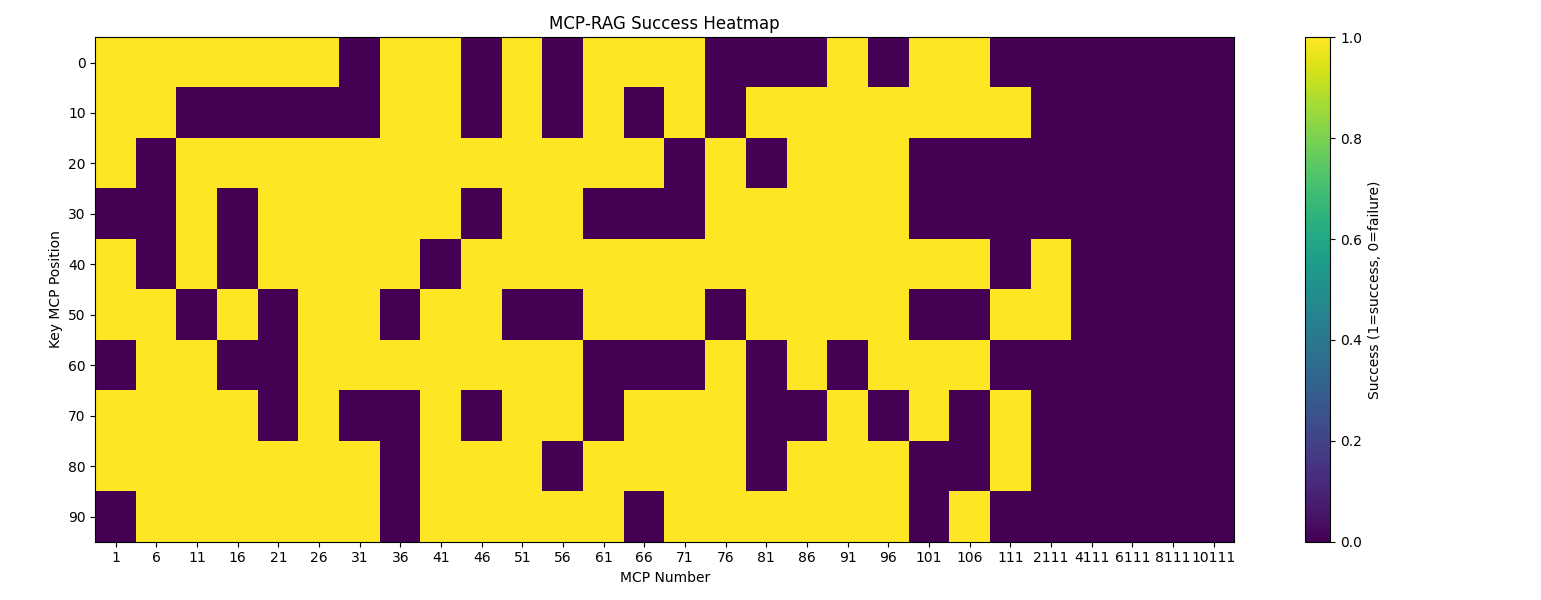} 
    \caption{This figure illustrates per-trial success across MCP positions from 1 to 11100, where yellow denotes successful selection and purple denotes failure.}
    \label{fig:process}
\end{figure}

\subsubsection{Results}
Figure3 plots selection accuracy and task success as $N$ increases. We observe a clear non-monotonic trend:
These results quantitatively confirm that while MCP‑RAG greatly mitigates prompt bloat and maintains high performance in small to moderate pools, its retrieval precision and overall throughput degrade as the tool registry scales to thousands of MCPs.```

\subsection{RAG-MCP}
\subsubsection{Setup}

We evaluated all methods in the web search subset of MCPBench \cite{MCPBench}, which we used as our heldout testbed. For each baseline, we perform 20 independent trials, and we deem a baseline successful if it produces more than 10 correct answers out of those 20. Within each trial, the model may engage in up to 10 rounds of interaction with the MCP servers in order to arrive at its final response.

To assess answer correctness in an automated and reproducible manner, we employ Deepseek‑v3 \cite{liu2024deepseek} as our evaluator. Because MCP servers require external network access—and can therefore be sensitive to latency or transient failures—we enforce a controlled network environment throughout all experiments, ensuring no requests fail due to connectivity issues. Finally, all trials are driven by qwen‑max‑0125 as our underlying base LLM.

\subsubsection{Baselines} 

We evaluate three selection strategies in our experiments:
\begin{enumerate}[1.]
  \item \textbf{Blank Conditioning}: Prompt the LLM with all $N$ MCP descriptions at once and ask it to choose the correct one.
  \item \textbf{Actual Match}: Pre-filter the candidate pool using simple keyword matching on the task description and MCP metadata, then prompt the model on this reduced set.
  \item \textbf{RAG‑MCP}: Employ our vector-index retriever to semantically rank all $N$ MCPs and inject only the top candidate’s schema into the LLM prompt for execution.
\end{enumerate}

\subsubsection{Metrics}
We evaluate performance using three key metrics for each baseline method:
\begin{itemize}
  \item \textbf{Accuracy (\%)}: Percentage of trials in which the model selected the ground‑truth MCP.
  \item \textbf{Avg Prompt Tokens}: Mean number of tokens consumed by the prompt, including injected MCP metadata.
  \item \textbf{Avg Completion Tokens}: Mean number of tokens generated by the model as its final output.
\end{itemize}
Judgment of the final answer is automated using a Llama-based verifier (“Llama as Judge”) to compare model outputs against ground truth.

\subsubsection{Results}

Table \ref{tab:baseline-comparison} summarizes the performance of the evaluated baseline methods, clearly demonstrating the effectiveness of MCP-RAG:

\begin{table}[h]
\centering
\renewcommand\arraystretch{1.2}
\begin{tabular}{lccc}
\hline
\textbf{Baseline} & \textbf{Accuracy (\%)} & \textbf{Avg Prompt Tokens} & \textbf{Avg Completion Tokens} \\
\hline
MCP-RAG       & 43.13      & 1084.00               & 78.14                    \\
Actual Match  & 18.20      & 1646.00               & 23.60                    \\
Blank         & 13.62      & 2133.84               & 162.25                   \\
\hline
\end{tabular}
\caption{Baseline performance comparison on accuracy and token usage}
\label{tab:baseline-comparison}
\end{table}

As the table shows, \textbf{MCP-RAG} achieves the highest accuracy at \textbf{43.13\%}, significantly outperforming the Actual Match and Blank Conditioning methods, which scored \textbf{18.20\%} and \textbf{13.62\%}, respectively. Furthermore, MCP-RAG notably reduces the average number of prompt tokens to \textbf{1084}, reflecting a substantial reduction compared to the other baselines, especially Blank Conditioning, which requires \textbf{2133.84} tokens. While MCP-RAG shows an increase in completion tokens (\textbf{78.14}) compared to Actual Match (\textbf{23.60}), this trade-off is beneficial as it correlates with a higher accuracy and overall task success rate.

\section{Analysis}

\subsubsection{Stress Test Analysis}
Figure 3 illustrates per-trial success across MCP positions from 1 to 11100, where yellow denotes successful selection and purple denotes failure. We observe that:
\begin{itemize}
  \item \textbf{High Early-Stage Success}: MCP positions below 30 exhibit predominantly yellow regions, indicating success rates above 90\% when the candidate pool is minimal.
  \item \textbf{Mid-Range Variability}: In the range of positions 31--70, clusters of purple emerge intermittently, reflecting lower accuracy as semantic overlap among MCP descriptions increases.
  \item \textbf{Performance Degradation at Scale}: Beyond position \textasciitilde100, purple dominates, signifying that retrieval precision diminishes when handling very large tool registries.
  \item \textbf{Residual Success Islands}: Occasional yellow patches at higher positions suggest that certain MCPs remain well-aligned to specific queries, providing robustness even in extensive pools.
\end{itemize}
These patterns confirm that while MCP-RAG effectively curbs prompt bloat and maintains high accuracy in small to moderate MCP pools, retrieval precision challenges arise as the total number of MCPs grows, motivating future work on hierarchical or adaptive retrieval mechanisms.

\subsection{Analysis of RAG‑MCP Results}
The superior performance of RAG‑MCP can be attributed to several factors:
\begin{itemize}
  \item \textbf{Focused Context Filtering}: By injecting only the single most relevant MCP schema, the model avoids the distraction caused by irrelevant tool descriptions, resulting in clearer decision boundaries.
  \item \textbf{Prompt Efficiency}: The dramatic reduction in prompt tokens allows the model to allocate more of its context window to reasoning about the task itself rather than parsing extraneous metadata.
  \item \textbf{Balanced Generation}: Although RAG‑MCP slightly increases completion token usage relative to Actual Match, this overhead reflects more thorough reasoning and verification steps, which correlate with higher accuracy.
\end{itemize}

Overall, these findings confirm that retrieval‑augmented selection of MCPs effectively tames prompt bloat and enhances an LLM’s tool‑selection reliability, making RAG‑MCP a compelling solution for scalable external tool integration.

\section{Conclusion}

We present \textbf{RAG‑MCP}, a simple yet powerful framework that tames large MCP toolsets by retrieving only the most relevant schema for each query. With focused retrieval, RAG‑MCP:
\begin{itemize}
  \item \textbf{Drastically reduces prompt size}, cutting token usage by over half compared to feeding all tools at once.
  \item \textbf{Boosts selection accuracy}, more than tripling the success rate of naïve and keyword‑based methods under heavy load.
  \item \textbf{Maintains extensibility}, since new MCPs can be indexed on–the–fly without retraining the model.
\end{itemize}

In essence, RAG‑MCP turns a sprawling library of hundreds or thousands of tools into a lean, on‑demand toolkit. Future work will refine retrieval at extreme scale—via hierarchical indexes or adaptive strategies—and explore multi‑tool workflows and real‑world agent deployments. RAG‑MCP lays the “golden core” for scalable, reliable LLM agents that wield vast external services with precision and efficiency. 

%
%

\bibliographystyle{splncs04}
\bibliography{reference}

\section{Acknowledgements}

We gratefully acknowledge Zhiling Luo, Xiaorong Shi, Xuanrui Lin, and Jinyang Gao for their seminal report “Evaluation Report on MCP Servers.” Their publicly released code framework and the accompanying WebSearch dataset formed the foundation on which our own work was developed.

\end{document}